\documentclass[10pt,twocolumn,letterpaper]{article}

\usepackage[pagenumbers]{cvpr} % To force page numbers, e.g. for an arXiv version

\definecolor{cvprblue}{rgb}{0.21,0.49,0.74}
\usepackage[pagebackref,breaklinks,colorlinks,allcolors=cvprblue]{hyperref}

\def\paperID{11} % *** Enter the Paper ID here
\def\confName{CVPR}
\def\confYear{2026}

\usepackage{amsmath}
\usepackage{graphicx}
\usepackage{amssymb}
\usepackage{wrapfig}

\usepackage{multirow}
\usepackage{multicol}
\usepackage{soul}
\title{Training-Free Layout-to-Image Generation with
Marginal Attention Constraints}

\author{Huancheng Chen\textsuperscript{1,2}, Jingtao Li\textsuperscript{1}, Weiming Zhuang\textsuperscript{1}, Haris Vikalo\textsuperscript{2}, Lingjuan Lyu\textsuperscript{1,*}\\
\textsuperscript{1}Sony AI, \textsuperscript{2}University of Texas at Austin \\
{\tt\small \{huanchengch, hvikalo\}@utexas.edu, \{jingtao.li, weiming.zhuang, lingjuan.lv\}@sony.com}
}

\begin{document}
\maketitle

\begin{abstract}
Recently, many text-to-image diffusion models have excelled at generating high-resolution images from text but struggle with precise control over spatial composition and object counting. To address these challenges, prior works have developed layout-to-image (L2I) approaches that incorporate layout instructions into text-to-image models. However, existing L2I methods typically require fine-tuning of pre-trained parameters or training additional control modules for diffusion models. In this work, we propose a training-free L2I approach, MAC (Marginal Attention Constrained Generation), which eliminates the need for additional modules or fine-tuning. Specifically, we use text–visual cross-attention feature maps to quantify inconsistencies between the layout of the generated images and the provided instructions, and then compute loss functions to optimize latent features during the diffusion reverse process. To enhance spatial controllability and mitigate semantic failures under complex layout instructions, we leverage pixel-to-pixel correlations in self-attention feature maps to align cross-attention maps and combine three loss functions constrained by boundary attention to update latent features. Comprehensive experimental results on both L2I and non-L2I pretrained diffusion models demonstrate that our method outperforms existing training-free L2I techniques, both quantitatively and qualitatively, in terms of image composition on the DrawBench and HRS benchmarks.
\end{abstract}  

\section{Introduction}
In recent years, text-to-image (T2I) diffusion models such as DALL·E \citep{Dall-E}, Imagen \citep{Imagen}, and Stable Diffusion \citep{LDM} have demonstrated an impressive ability to generate high-resolution images from textual input.\renewcommand\thefootnote{}
\footnote{*Corresponding Author: Lingjuan Lyu}
\renewcommand\thefootnote{\arabic{footnote}}These models derive their ability to unify textual and visual latent spaces from supervised training on large-scale datasets \citep{clip,laion} of text–image pairs sourced from the internet. While such generative models have achieved remarkable success in various downstream tasks, textual input alone cannot accurately specify spatial composition or the precise locations of different concepts within generated images. 

\begin{figure*}[t]
\begin{center}
\includegraphics[width= 1.0\linewidth]{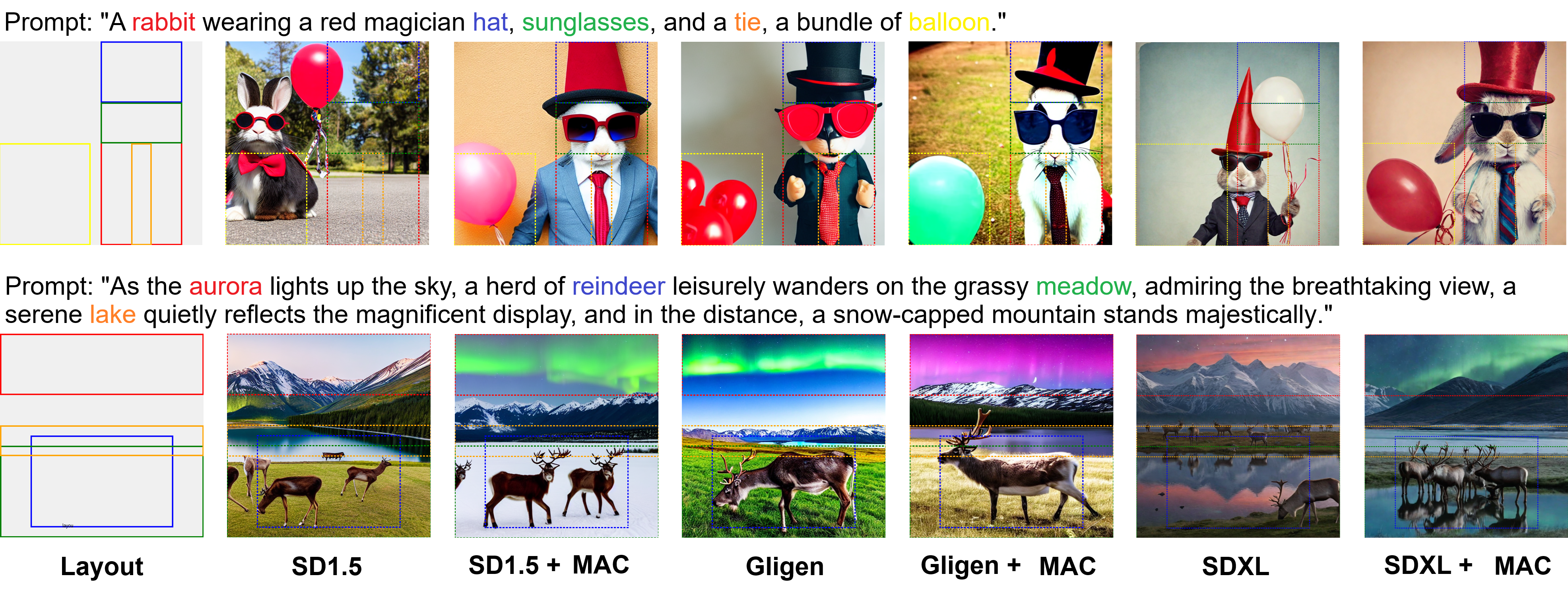}
\end{center}
   \caption{An illustration of training-free layout-to-image (L2I) generation using various diffusion models. Text prompts and layout information, where specific concepts are localized within corresponding bounding boxes, are provided as input to pre-trained diffusion models to generate images that align with the instructions. Our method, MAC, is capable of guiding non-L2I diffusion models such as SD \citep{LDM} and SDXL \citep{sdxl} to generate images based on layout instructions, while also enhancing L2I models such as GLIGEN \citep{gligen} to achieve improved spatial control.} 
\label{plugandplay}
\end{figure*}

To address this deficiency, numerous studies \citep{reco, geodiffusion, gligen, layout, controlnet} have explored layout-to-image (L2I) generation approaches that enable the localization of concept positions in prompts using various forms of layout instructions, such as semantic masks, bounding boxes, or sketches. By incorporating additional layout information, text-to-image (T2I) models can achieve more precise spatial controllability and generate datasets with specific ground-truth labels for data augmentation in supervised training \citep{gligen, reco}. However, these L2I approaches require adapting pre-trained T2I models with additional control modules that must be fine-tuned on large datasets containing paired images and layout annotations, such as COCO \citep{coco}. 

The significant data and computational requirements for model fine-tuning restrict the applicability of such approaches in many data-scarce or resource-constrained scenarios. To this end, a series of methods for layout-to-image generation in a zero-shot manner, i.e., requiring no additional supervised training, have been proposed \citep{loco, boxdiff, rnb, layout_guidance, anr}. These techniques leverage the cross-attention maps extracted from the U-Net \citep{unet} in diffusion models to quantify the discrepancy between a synthetic image—sampled from the initialized latent feature $z_t$ based on the input text prompt—and the target layout instructions; the subsequent iterative update of the latent feature $z_t$ helps reduce this discrepancy. However, it has been observed that cross-attention maps typically exhibit high responses in the central regions of concepts while assigning negligible scores to their edges \citep{rnb, loco, anr}. Consequently, the generated concepts often appear larger than or misaligned with the specified bounding boxes, which diminishes the accuracy of layout control and leads to unsatisfactory image generation. Additionally, existing approaches often suffer from overlapping cross-attention maps when the bounding boxes for multiple objects of the same concept are closely spaced with minimal gaps, leading to inaccuracies in object counts relative to the input prompt.

To improve spatial controllability and address semantic failures, we propose MAC, which stands for \underline{M}arginal \underline{A}ttention \underline{C}onstrained Generation, a novel training-free L2I approach. Specifically, we introduce two key design principles: (1) leveraging pixel-to-pixel correlations from visual self-attention maps to align coarse-grained cross-attention maps, and (2) proposing a boundary attention constraint to address challenges related to size misalignment and inaccurate object counting. Self-attention maps capture pixel-to-pixel correlations in visual features, which can be used to filter noisy cross-attention maps and enhance the edges of concepts with low attention scores. Meanwhile, the boundary attention constraint ensures that the cross-attention for each individual object remains within the boundaries of its corresponding bounding box, while promoting the separation of cross-attention maps for multiple objects within the same concept. Our comprehensive experimental results demonstrate that the proposed method outperforms existing training-free L2I methods, both quantitatively and qualitatively, in terms of image composition—specifically \emph{spatial relationships}, \emph{size}, \emph{color}, and \emph{object counting}—on the \textbf{DrawBench} \citep{Imagen} and \textbf{HRS} \citep{hrs} benchmarks. The main contributions of this paper can be summarized as follows:
\begin{itemize}[leftmargin=0.35cm]
\item We investigate semantic failures in training-free L2I generation under complex layout inputs, with a particular focus on overlapping cross-attention maps that lead to inaccurate object counting.
\item We propose a novel method, MAC, which advances L2I generation performance by incorporating self-attention–based enhancement to filter noisy cross-attention maps and boundary attention constraints to prevent cross-attention overlap.
\item We conduct comprehensive experiments comparing our method with existing L2I approaches. Both quantitative and qualitative results demonstrate that our method achieves state-of-the-art performance among training-free L2I techniques.
\end{itemize}

\section{Related Work}
\subsection{Text-To-Image Diffusion Models}
Recently, diffusion models \citep{ddpm, ddim, sde} have achieved remarkable success in image generation, demonstrating greater stability and controllability compared to GANs \citep{gan}. Essentially, a diffusion model learns the reverse dynamics of a forward noising process that gradually maps natural image-like data to a simple prior distribution (e.g., Gaussian). The seminal work LDM \citep{LDM} facilitates high-resolution image synthesis by performing forward and reverse diffusion processes in the latent space rather than the pixel space, thereby addressing issues of low inference speed and high training costs.  In this framework, a U-Net \citep{unet} is trained as the denoiser $\boldsymbol{\epsilon}_{\theta}$ to predict the injected noise $\boldsymbol{\epsilon}$ from the perturbed latent variable $\mathbf{z}_{t}$, conditioned on the auxiliary input $\mathbf{c}$, by minimizing the following objective:
\begin{equation}
    L(\theta) = 
    \mathbb{E}_{\mathbf{z}_{0},\,\boldsymbol{\epsilon}\sim\mathcal{N}(\mathbf{0},\mathbf{I}),\,t}
    \left[
    \left\|
    \boldsymbol{\epsilon}
    -
    \boldsymbol{\epsilon}_{\theta}(\mathbf{z}_{t}, \mathbf{c}, t)
    \right\|^{2}
    \right].
\end{equation}

Building on the foundations of conditional generation with diffusion models, DALL·E \citep{Dall-E} advances text-to-image (T2I) generation by utilizing the powerful image--language encoder CLIP \citep{clip} to transform text prompts into conditioning embeddings $\mathbf{c}$ that guide the reverse diffusion process. Imagen \citep{Imagen} and Stable Diffusion \citep{LDM} further extend this paradigm by improving diffusion sampling techniques, enabling more photorealistic and detailed image generation. However, relying solely on text prompts limits the ability to precisely control spatial composition, necessitating additional inputs in the form of layout instructions.

\subsection{Layout-To-Image Generation}
In addition to text prompts, layout-to-image (L2I) models require auxiliary layout instructions as input, \emph{i.e.}, semantic masks or bounding boxes. Several approaches aim to map layout instructions into the condition embedding space of diffusion models through supervised fine-tuning on datasets consisting of paired layout and image data.
These training-based approaches either integrate bounding boxes into the text prompts \citep{reco, geodiffusion, layout, avrahami2023spatext} and fine-tune the text encoder, or incorporate additional modules or adapters alongside the text encoder to enhance the understanding of layout instructions \citep{gligen, controlnet}. However, this line of L2I approaches requires additional image--annotation paired datasets \citep{cheng2024hico,zhang2025creatilayout} to fine-tune the fusion module, demanding substantial human effort for data labeling.

Another line of training-free approaches aims to address the computational and data challenges associated with supervised L2I methods. Inspired by the connection between spatial layouts of generated images and the cross-attention maps observed in \citep{prompt}, DenseDiff \citep{dense} guides the placement of objects in targeted positions by manipulating the cross-attention maps in diffusion models. Follow-up studies such as A\&E \citep{attend-and-excite} and BoxDiff \citep{boxdiff} propose optimizing the latent variable $z_{t}$ by maximizing the supremum of cross-attention scores located in specific regions during the reverse process of diffusion models. A\&R \citep{anr} incorporates self-attention maps into the objective function to penalize objects that appear outside the designated boxes.
Recent training-free methods further improve spatial fidelity by revisiting attention-based guidance or addressing complex multi-object layouts, such as WinWinLay \citep{li2025winwinlay} and CSG \citep{liu2024csg}.

In contrast to previous studies that utilized the supremum, layout-guidance \citep{layout} uses the average of cross-attention scores in the objective function to update the latent variable $z_{t}$. Due to the typically coarse-grained and noisy nature of raw cross-attention maps, R\&B \citep{rnb} employs the Sobel Operator \citep{edge} to detect edges within these maps and select candidate boxes that encompass all edges for loss calculation. To address the additional computation and slower inference speed introduced by edge detection, LoCo \citep{loco} directly leverages start-of-text tokens (SoT), end-of-text tokens (EoT), and self-attention maps to enhance cross-attention maps. B2B \citep{taghipour2024box} incorporates object generation and attribute binding to improve the controllability. 
However, none of the aforementioned approaches incorporates marginal attention constraints, leading to failures in object counting and precise spatial controllability.

\section{Methodology}
\label{method}
\subsection{Training-Free Layout-To-Image Generation}
Training-free layout-to-image generation aims to synthesize images from input text prompts and corresponding layout instructions using a pre-trained diffusion model, without requiring additional parameter training or auxiliary modules. Consider a text prompt $\mathbf{p} = \{\mathbf{p}_{1}, \dots, \mathbf{p}_{n}\}$ and a set of bounding boxes $\mathcal{B} = \{\mathcal{B}_{i} \mid i \in \mathcal{I}, \ \mathcal{I} \subseteq [n]\}$. Here, $\mathcal{B}_{i} = \{\mathbf{b}^{(i)}_{1}, \dots, \mathbf{b}^{(i)}_{N_{i}}\}$ denotes the set of bounding boxes associated with phrase $\mathbf{p}_{i}$, where each bounding box is represented by a pair of top-left and bottom-right coordinates $(x_{1}, y_{1}, x_{2}, y_{2})$, specifying the spatial locations of $N_{i}$ objects described by $\mathbf{p}_{i}$. The generated images are expected to closely align with the text prompts while remaining consistent with the layout constraints defined by the bounding boxes $\mathcal{B}$. An illustration of training-free layout-to-image generation using bounding boxes as layout instructions is provided in Figure~\ref{plugandplay}.

\begin{figure*}[t]
\begin{center}
\includegraphics[width= 1.0\linewidth]{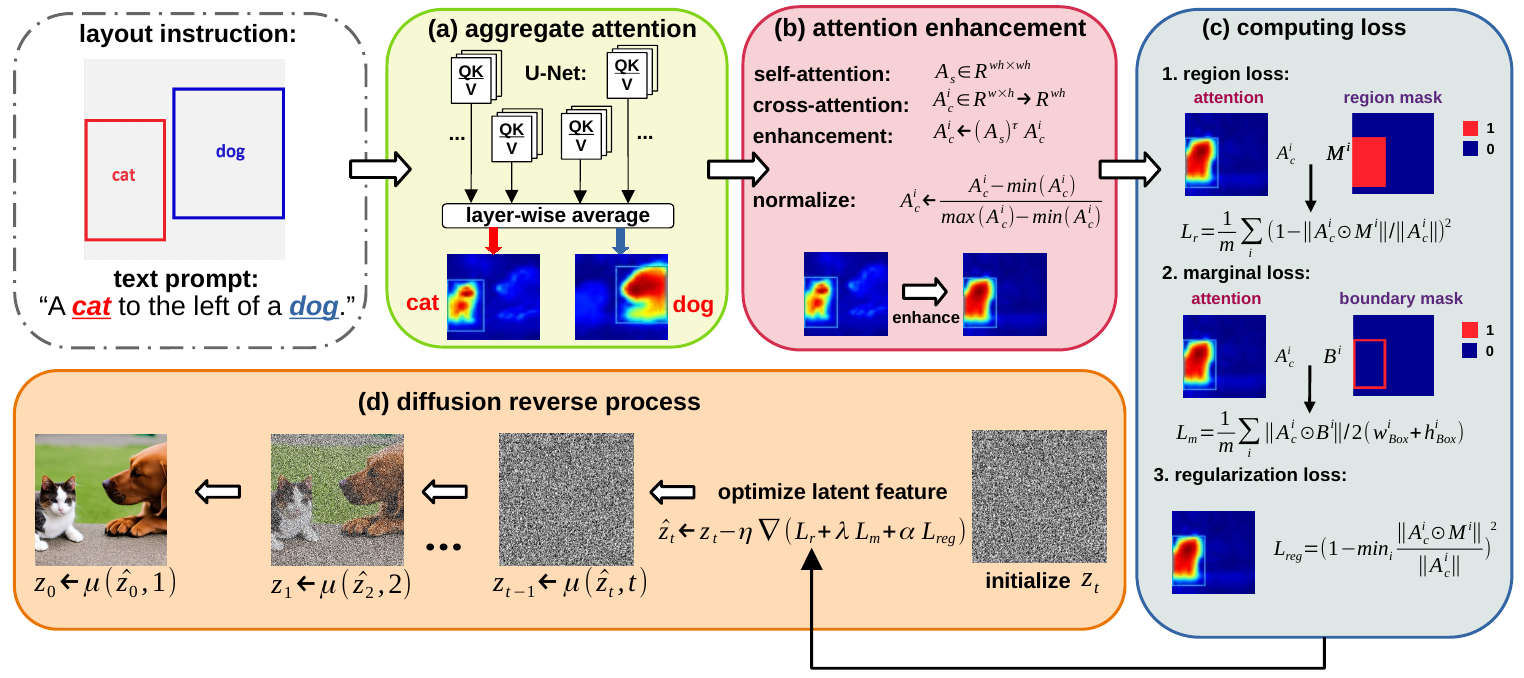}
\end{center}
   \caption{The overall framework of MAC for training-free L2I generation consists of four steps: (a) aggregating cross-attention and self-attention maps obtained from the language-vision fusion blocks by layer-wise averaging; (b) enhancing cross-attention maps with self-attention maps; (c) computing three losses and combining them with coefficients; (d) optimizing $z_{t}$ by gradient descent.} 
\label{bacon}
\end{figure*}

\subsection{Cross-Attention and Self-Attention Maps}
In text-to-image (T2I) diffusion models, a text prompt $\mathbf{p}$ is encoded into a sequence of text tokens $\mathbf{e} = f_{\text{CLIP}}(\mathbf{p}) \in \mathbb{R}^{n \times d_{e}}$ using a pre-trained CLIP encoder. These tokens are subsequently fed into the cross-attention layers, where they are fused with the visual embeddings represented by the latent feature $\mathbf{z}_{t}$. Specifically, at layer $l$, the visual and text embeddings are projected into the \emph{query} matrix $\mathbf{Q}_{z}^{l} \in \mathbb{R}^{hw \times d}$ and the \emph{key} matrix $\mathbf{K}_{e}^{l} \in \mathbb{R}^{n \times d}$, respectively, enabling the computation of cross-attention maps as
\begin{equation}
    \mathbf{A}_{c}^{l} =
    \text{softmax}\!\left(
    \frac{\mathbf{Q}_{z}^{l}(\mathbf{K}_{e}^{l})^{\top}}{\sqrt{d}}
    \right)
    \in [0,1]^{hw \times n},
\end{equation}
where $n$ denotes the length of the text prompt (including SoT and EoT tokens), $d_{e}$ and $d$ denote the dimensions of the text and visual embeddings, respectively, and $h$ and $w$ represent the height and width of the visual feature maps. Similarly, the visual embeddings are projected into the \emph{key} matrix $\mathbf{K}_{z}^{l} \in \mathbb{R}^{hw \times d}$, allowing the computation of self-attention maps
\begin{equation}
    \mathbf{A}_{s}^{l} =
    \text{softmax}\!\left(
    \frac{\mathbf{Q}_{z}^{l}(\mathbf{K}_{z}^{l})^{\top}}{\sqrt{d}}
    \right)
    \in [0,1]^{hw \times hw},
\end{equation}
which capture pixel-to-pixel similarities in the visual features. The cross-attention and self-attention maps extracted from $L$ different attention layers are then aggregated as
\begin{equation}
    \mathbf{A}_{c} = \frac{1}{L}\sum_{l=1}^{L}\mathbf{A}_{c}^{l},
    \quad
    \mathbf{A}_{s} = \frac{1}{L}\sum_{l=1}^{L}\mathbf{A}_{s}^{l}.
\end{equation}

\subsection{Marginal Attention Constrained Guidance}
While prior studies \citep{rnb,layout_guidance} demonstrated effectiveness of backward guidance, a procedure where cross-attention scores are used to form the loss utilized for optimizing the latent feature $z_{t}$ during the diffusion reverse process, the designed loss functions fail to address issues related to counting and semantic failures arising due to the coarse-grained nature of the cross-attention maps. In this section, we discuss the effect of attention enhancement and present three loss functions to address the problems encountered by previous schemes.
\\

\noindent \textbf{Attention Enhancement.} As depicted in the pink block in Figure \ref{bacon}, the raw cross-attention map for ``cat" is coarse-grained, with attention dispersed across multiple parts of the object. It also shows notable scores for ``dog" in the cross-attention map for the ``cat", likely due to the high correlation between ``cat" and ``dog" within the pretrained model, which causes inconsistency in the loss computation. Inspired by the previous studies \citep{ma2023diffusionseg,nguyen2024dataset} that aim to build synthetic datasets by generating grounded labels using cross-attention, we leverage the global information contained in self-attention maps to enhance the cross-attention maps, 
\begin{equation}
    \mathbf{A}_{c}^{(i)} = \left(\mathbf{A}_{s}\right)^{\tau} \cdot \mathbf{A}_{c}^{(i)} \in [0,1]^{wh},
\end{equation}
where $i$ denotes the index of the phrase $\mathbf{p}_{i}$, and $\tau$ is a power coefficient for adjusting the magnitude of enhancement (set to $1$ by default). For dimensional consistency, we normalize and reshape $\mathbf{A}_{c}^{(i)}$ according to
\begin{equation}
    \mathbf{A}_{c}^{(i)} = \textbf{reshape}\left(\frac{\mathbf{A}_{c}^{(i)} - \min(\mathbf{A}_{c}^{(i)})}{\max(\mathbf{A}_{c}^{(i)}) - \min(\mathbf{A}_{c}^{(i)})}\right) \in [0,1]^{w \times h}.
\end{equation}
Essentially, the self-attention map $\mathbf{A}_{s}$ captures the pairwise correlations between pixels, allowing salient parts in the raw cross-attention maps to propagate to the most relevant regions. Using these enhanced cross-attention maps, we compute three losses for improving the L2I generation.
\\

\noindent \textbf{Region-Attention Loss.}
Similar to the prior studies \citep{rnb, anr, layout_guidance}, the key idea of region-attention loss is to measure the proportion of cross-attention map $\mathbf{A}_{c}^{(i)}$ outside bounding boxes $\mathcal{B}_{i} \in \mathcal{B}$ and compute the average of these proportions for normalization as
\begin{equation}
 \mathbf{M}^{(i)}[x,y] = \left\{
    \begin{aligned}
        & \; 1, \text{ if } (x,y) \text{ inside } \mathcal{B}_{i} \\
        & \; 0, \text{ if } (x,y) \text{ outside } \mathcal{B}_{i}\\
    \end{aligned}
    \right.,
\end{equation}
\begin{equation} 
    L_{\text{r}} = \frac{1}{|\mathcal{B}|}\sum_{i \in \mathcal{I}}\left(1 - \frac{\mathbf{A}_{c}^{(i)}\odot \mathbf{M}^{(i)}}{\sum_{x,y}\mathbf{A}_{c}^{(i)}[x,y]}\right)^{2}, 
\end{equation}
where $\mathbf{A}_{c}^{(i)}[x,y]$ and $\mathbf{M}^{(i)}[x,y]$ are the (x,y)-th element in $\mathbf{A}_{c}^{(i)}$ and $\mathbf{M}^{(i)}$, respectively. However, relying solely on region-attention loss $L_{\text{r}}$ cannot ensure high-quality generation in scenarios where multiple objects constrained by the same $\mathcal{B}_{i}$ have overlapping cross-attention scores, resulting in incorrect counting in the generated image.
\\

\noindent \textbf{Marginal-Attention Loss.}
As previously mentioned, adjacent bounding boxes for multiple objects described by the same phrase often lead to incorrect counting in the L2I generation due to the interference between cross-attention scores. However, the region-attention loss may remain small as long as these cross-attention scores are inside the target bounding boxes. We propose a marginal-attention loss that aims to isolate adjacent cross-attention maps corresponding to different objects,
\begin{equation}
    \mathbf{B}^{(i)}[x,y] = \left\{
    \begin{aligned}
        & \; 1, \text{ if } (x,y) \text{ on } \mathcal{B}_{i} \\
        & \; 0, \text{ if } (x,y) \text{ not on } \mathcal{B}_{i}\\
    \end{aligned}
    \right.,
\end{equation}
\begin{equation}
    L_{\text{m}} = \frac{1}{|\mathcal{B}|}\sum_{i \in \mathcal{I}}\frac{\mathbf{A}_{c}^{(i)}\odot \mathbf{B}^{(i)}}{2(w_{\text{Box}} + h_{\text{Box}})},
\end{equation}
where $w_{\text{Box}} = \sum_{k=1}^{N_{i}}w_{k}^{(i)}$, $h_{\text{Box}} = \sum_{k=1}^{N_{i}}h_{k}^{(i)}$, and $w_{k}^{(i)}$ and $h_{k}^{(i)}$ denote the width and height of the bounding box $\mathbf{b}_{k}^{(i)}$, respectively. With the proposed marginal-attention loss, we can force $z_{t}$ to generate cross-attention maps without overlap between multiple objects and thus help improve the accuracy of counting.
 \\
 
\noindent \textbf{Regularization Loss.}
While the region-attention and marginal-attention losses generally help localize generated objects within the target bounding boxes, these losses may be minimal when an incorrect number of objects has cross-attention scores completely contained within the bounding boxes (i.e., experiencing no boundary crossing). To prevent bad generation in such scenarios, we propose a regularized loss that ensures the assigned objects do not have a void cross-attention map
\begin{equation}
  L_{\text{reg}} = \frac{1}{|\mathcal{B}|}\sum_{i \in \mathcal{I}}\left(1 - \min_{k}\frac{\mathbf{A}_{c}^{(i)}\odot \mathbf{M}^{(i,k)}}{\sum_{x,y}\mathbf{A}_{c}^{(i)}[x,y]}\right)^{2}, 
\end{equation}
where $\mathbf{M}^{(i,k)}$ is a subset of $\mathbf{M}^{(i)}$ indicating the location of the $k$-th object in $\mathcal{B}_{i}$.
\\

\noindent \textbf{Latent Feature Optimization.}
At each sampling time step in the reverse process, $L_{\text{mac}}$ can be computed as the weighted summation of $L_{\text{r}}$, $L_{\text{m}}$ and $L_{\text{reg}}$,
\begin{equation}
\label{loss_bacon}
    L_{\text{mac}} = L_{\text{r}} + \lambda L_{\text{m}} + \alpha L_{\text{reg}},
\end{equation}
where $\lambda$ and $\alpha$ control the intervention strength of $L_{\text{m}}$ and $L_{\text{reg}}$, respectively. We compute and use the gradient of $L_{\text{mac}}$ to update $z_{t}$ as
$
    \hat{z_{t}} \xleftarrow{} z_{t} - \eta \nabla L_{\text{mac}},
$
where parameter $\eta$ controls the size of the updates. After $t_{\text{optim}}$ iterations or an early stop that happens when $L_{\text{mac}}$ becomes smaller than a predetermined threshold, the optimized latent feature $\hat{z_{t}}$ is forwarded to the U-Net for predicting the latent feature $z_{t-1}$ at the previous time step.

\section{Experiments}
\label{exp}
\subsection{Experimental Setup}
\textbf{Datasets and base models.}
Following the strategy in the prior studies \citep{rnb,anr}, we perform experiments on a subset of two widely used benchmarks \textbf{HRS} \citep{hrs} and \textbf{Drawbench} \citep{Imagen}. Specifically, we use four tracks from the HRS benchmark -- \textbf{spatial relationship}, \textbf{size}, \textbf{color}, and \textbf{object counting} -- which include $1002$, $501$, $501$, and $3000$ text prompts, respectively. These prompts, along with the corresponding layout instructions generated by ChatGPT-4 in \citep{anr}, are used to evaluate the performance of our method and baseline approaches both quantitatively and qualitatively. Similarly, $39$ text prompts from Drawbench involving \textbf{spatial} and \textbf{counting} specifications are also used in the evaluation. We conduct experiments with the widely-used Stable Diffusion (SD) 1.5 \citep{LDM} as the base model and perform an ablation study with SD-XL  \citep{sdxl} and the fine-tuned L2I model, Gligen \citep{gligen}. The sampling time step is set to $50$, with the classifier-free guidance weight set to $7.5$. Optimization is applied only to the latent feature $z_{t}$ during the first 10 sampling steps, and $\eta$ is set to 70. Unless specified otherwise, the hyperparameters $\lambda$ and $\alpha$ are both set to $0.5$.
\\

\noindent \textbf{Baselines and metrics.} We compare our proposed scheme, MAC, to six state-of-the-art methods: A\&E \citep{attend-and-excite}, BoxDiff \citep{boxdiff}, Layout-Guidance \citep{layout_guidance}, A\&R \citep{anr}, R\&B \citep{rnb}, and LoCo \citep{loco}. To quantitatively assess our method against these baselines, we use the state-of-the-art object detector, Ground-DINO \citep{grounddino}, to detect objects in the synthetic images and predict bounding boxes. These predicted boxes are then compared with the ground truth. We compute precision, recall, and F1 metrics to comprehensively evaluate MAC’s performance in object counting. By comparing the area and centroid of the predicted and ground-truth boxes, we compute the accuracy of object size and spatial relationships. Additionally, Ground-DINO predicts the color of detected objects, which we use to compute color accuracy. The details of the metric computations can be found in the Appendix B.\\

\begin{figure*}[t] 
    \centering
	  \subfloat[Layout]{
       \includegraphics[width=0.12\linewidth]{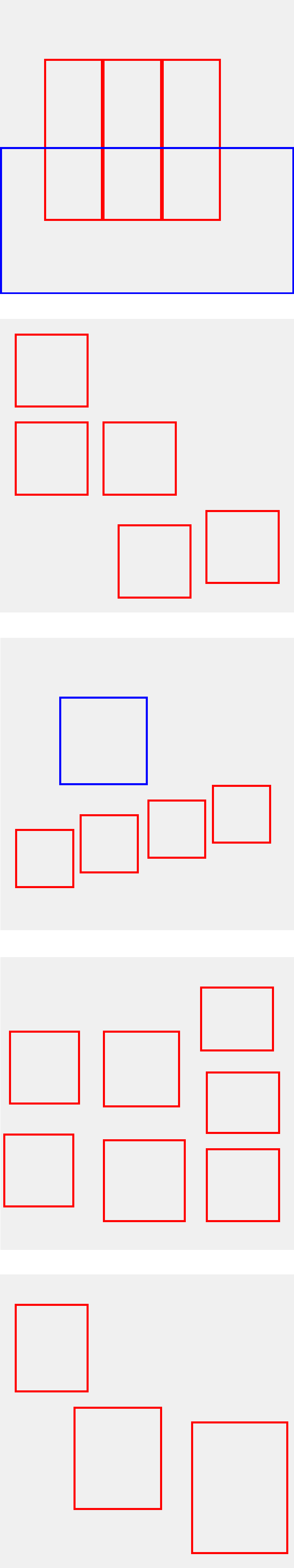}}
       \subfloat[Prompt]{
        \includegraphics[width=0.12\linewidth]{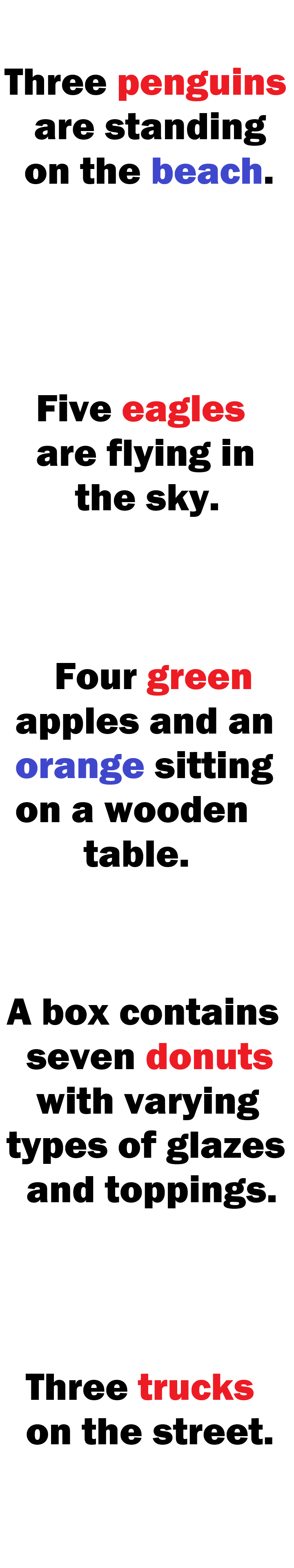}}
        \subfloat[Layout]{
        \includegraphics[width=0.12\linewidth]{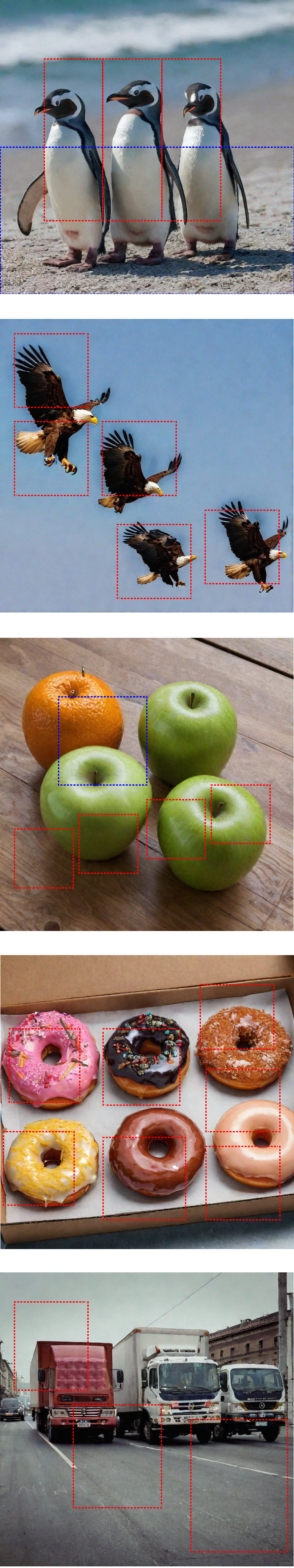}}
         \subfloat[A\&R]{
        \includegraphics[width=0.12\linewidth]{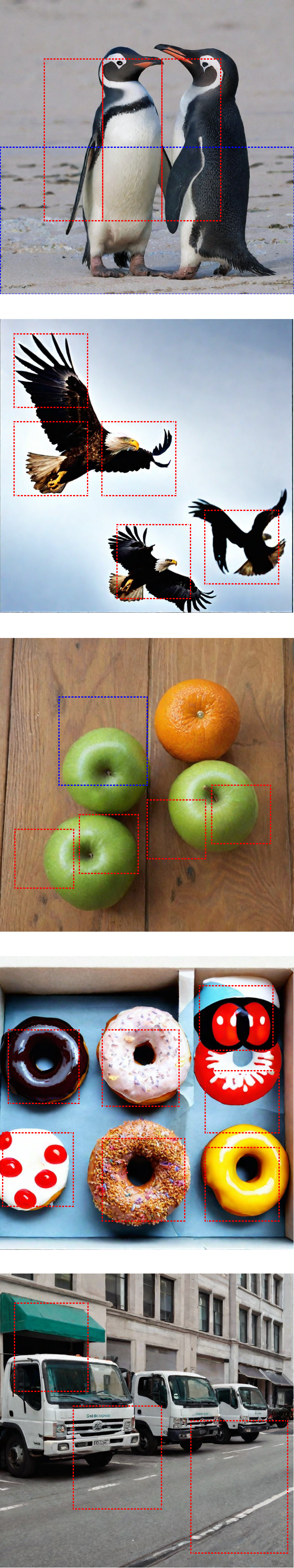}}
        \subfloat[R\&B]{
        \includegraphics[width=0.12\linewidth]{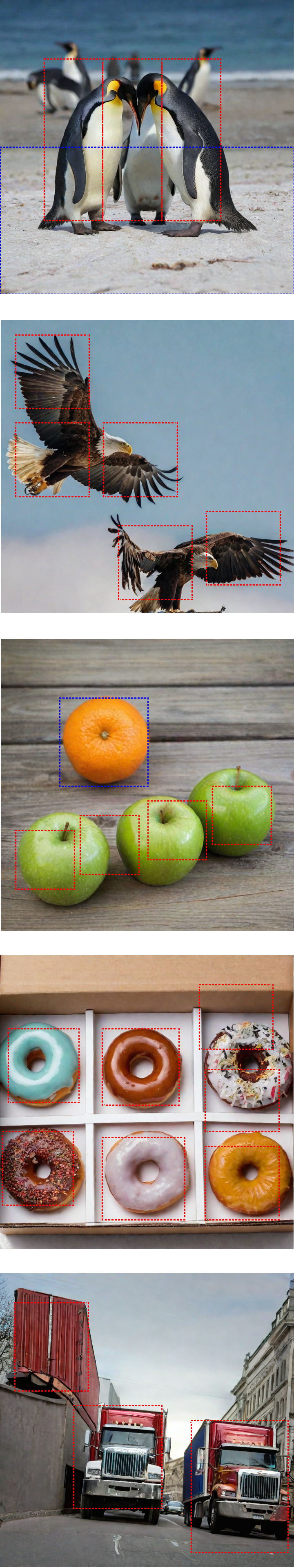}}
         \subfloat[MAC]{
        \includegraphics[width=0.12\linewidth]{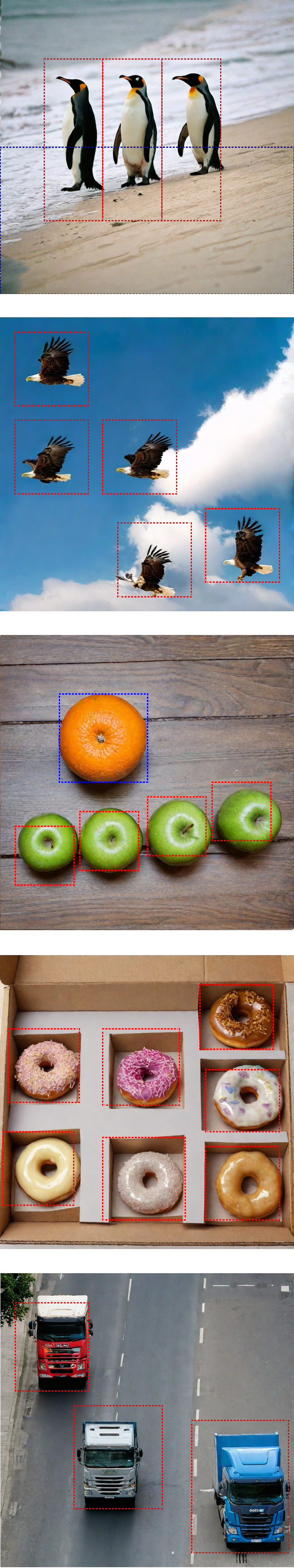}}

	\caption{Visual comparison with different training-free L2I schemes based on Stable Diffusion XL \citep{sdxl}. The layout input and text prompt are shown in the first and second columns, respectively. The layout instructions are also annotated on the generated images with dashed boxes. Our method, MAC, significantly outperforms prior schemes in terms of spatial control and counting accuracy.}
 \label{visual}
\end{figure*}

\vspace{-0.15 in}
\subsection{Visual Comparisons}
In Figure \ref{visual}, we present visual comparisons between MAC and several competing L2I schemes -- Layout, A\&R, and R\&B -- deployed on SD-XL \citep{sdxl}. To thoroughly evaluate the performance of these schemes, we use several complex inputs featuring multiple objects distributed in random positions, which makes it more challenging to generate images with precise object counts. As shown in the figure, MAC demonstrates significant improvement over the baselines in terms of (1) better alignment within the bounding boxes, and (2) more precise object counting. Although the prior works typically generate objects within the corresponding bounding boxes, most of these objects are only partially inside the boxes or even appear in the bounding boxes of other objects due to coarse-grained cross-attention maps, as we discussed earlier. For example, Layout \citep{layout} and A\&R \citep{anr} fail to generate penguins (1st row) within the red bounding boxes and an orange (3rd row) within the blue bounding box. Additionally, none of these prior schemes can generate the correct number of objects according to the prompts and bounding boxes specifications.

\begin{figure}[t] 
    \centering
	  \subfloat[A\&R]{
       \includegraphics[width=0.3\linewidth]{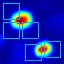}}
        \subfloat[R\&B]{
        \includegraphics[width=0.3\linewidth]{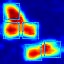}}
	  \subfloat[MAC]{
        \includegraphics[width=0.3\linewidth]{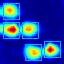}}
	\caption{Visualization of cross-attention maps with the prompt: \emph{Five eagles are flying in the sky.}}
\label{cross_attn} 
\end{figure}

To better understand the miscounting failures, we visualize the cross-attention maps obtained from these schemes as shown in Figure~\ref{cross_attn}. Since the cross-attention maps for objects described by the same phrase in the prompt are not independent, existing training-free L2I schemes, which aim to encourage cross-attention within the corresponding boxes, often produce overlapping cross-attention maps. For instance, in the R\&B scheme, the cross-attention scores are concentrated within three boxes, representing a single eagle, even though three eagles are meant to be depicted. However, marginal-attention loss proposed in MAC can isolate these cross-attention scores to ensure they depict independent objects in the generated images.

\begin{table*}[t]
\caption{Quantitative comparisons with baseline schemes. The precision (\%), recall (\%) and F1 score (\%) with respect to object counting are reported in the table. }
\centering	
\small
\begin{tabular}{lccccccc}
\bottomrule[1pt]
\label{table1}
\multirow{2}{*}{Method} & \multicolumn{3}{c}{DrawBench} & \multicolumn{3}{c}{HRS-Bench} & \multirow{2}{*}{Inference (s)}\\
\cline{2-7}
 &Precision & Recall & F1 & Precision & Recall & F1 \\ 
 \hline
SD1.5 &76.51 &70.58 &71.35 &  71.86 &52.19 &58.31 & 14\\
+ A\&E \citep{attend-and-excite}   &75.52 &66.17 &76.27 &73.10 &54.79 &60.47 &42  \\
+ BoxDiff \citep{boxdiff} &86.36 &77.94 &86.17 &73.57 &63.33 &68.07  &42 \\
+ LoCo \citep{loco} &86.04 &54.41 &66.66 & 85.61 &59.55 & 71.23 & 39\\
+ Layout \citep{layout_guidance} &85.93 &80.88 &83.33 &87.25 &63.10 &73.24 &\textbf{37} \\
+ A\&R \citep{anr} &86.00 &63.23 &72.88 &87.93 &56.69 & 68.94 & 45\\
+ R\&B \citep{rnb} &87.50 &82.35 &84.84 &85.73 &68.53 &76.17  & 84\\
% + CSG \citep{liu2024csg} & & & & & &  & \\
% + WinWinLay \citep{li2025winwinlay} & & & & & &  & \\
+ \textbf{MAC (ours)} 
  &\textbf{93.84} &\textbf{89.71} &\textbf{91.72 }&\textbf{89.51} &\textbf{68.92} &\textbf{77.88}& 41\\
\toprule[1pt]
\end{tabular}
\end{table*}

\begin{table*}[t]
\caption{Quantitative comparisons with baseline schemes. The evaluated accuracy (\%) with respect to spatial relationship, size relationship and color are reported in the table.}
\centering
\small
\begin{tabular}{cccccccccc}
\bottomrule[1pt]
\label{table2}
Benchmarks & Metrics  & SD   & A\&E & BoxDiff & LoCo &Layout &A\&R  &R\&B  & \textbf{MAC} \\
\hline
DrawBench&Spatial &12.50 &15.00 &25.00 & 40.00&45.00 &40.00 & 55.00&  \textbf{60.00 }\\
\hline
\multirow{3}{*}{HRS-Bench} &Size &11.23  & 14.77  &12.77 & 14.56 &14.37 &12.17 &16.17  &\textbf{16.96} \\
&Color & 13.01 &18.27 &34.69 &37.88&31.18 &30.15 &36.36 &\textbf{39.74} \\
&Spatial &10.80 &17.56  &19.76 & 37.42 &33.73 &25.94 &47.80  &\textbf{50.39} \\
\toprule[1pt]
\end{tabular}
\end{table*}

\subsection{Quantitative Results}
We validate the improvement in spatial controllability and object counting of the proposed MAC framework as compared to existing state-of-the-art training-free L2I approaches through quantitative experiments. As shown in Tables \ref{table2} and \ref{table1}, MAC outperforms state-of-the-art baselines on both DrawBench and HRS benchmarks. Notably, MAC demonstrates significant improvements in object counting across three metrics: precision, recall, and F1 score. On DrawBench, MAC outperforms the best baseline, R\&B, by $6.34\%$, $7.36\%$, and $6.88\%$ in precision, recall, and F1 score, respectively. On the larger HRS benchmark, MAC consistently achieves the best performance among all L2I schemes, with improvements of $3.78\%$, $0.39\%$, and $1.71\%$ in precision, recall, and F1 score, respectively. Furthermore, MAC surpasses R\&B in the spatial relationship metric on both DrawBench and HRS benchmark, with improvement of $5\%$ and $2.59\%$. Additionally, MAC achieves the best performance in color and size metrics, though its advantage over R\&B in these areas is relatively small.

Nevertheless, R\&B requires applying Sobel operator \citep{sobel} to detect the edge of cross-attention map for creating a minimum bounding rectangle (MBR), which is then used to compute region-aware loss. Therefore, R\&B needs more computation in the process of optimizing latent feature $z_{t}$, which leads to longer inference time. As shown in Table.~\ref{table2}, R\&B needs more than double time on average for generating one image compared to MAC, even though R\&B has comparable performance to MAC.

\begin{table*}[t]
\caption{Zero-shot schemes can be integrated into existing fully-supervised layout-to-image schemes, \emph{e.g.}, GLIGEN \citep{gligen}. We compare the improvement of GLIGEN augmented with MAC vs. the competing schemes.}
\centering	
\small
\begin{tabular}{lccccccc}
\bottomrule[1pt]
\label{table3}
\multirow{2}{*}{Method} & \multicolumn{2}{c}{DrawBench} & & \multicolumn{4}{c}{HRS-Bench}\\
\cline{2-3}
\cline{5-8}
 &Counting (F1)  & Spatial &  & Counting (F1) & Spatial & Size & Color \\ 
 \hline
GLIGEN &  77.03 & 40.00 & &68.32 &44.81 &35.31 &30.55 \\
+ LoCo &76.61 &40.00 & & 71.89 &49.48 & 38.33 &37.97 \\
+ Layout &80.59 & 55.00 & &81.85 & 50.68 &29.64 &35.05 \\
+ A\&R &76.27  &55.00 & &77.66 &52.17 &30.72 & 38.74\\
+ R\&B &87.59 & \textbf{65.00} & &79.10 &53.88 & 39.72 & 40.59 \\
% + CSG & &  & & & &  &  \\
% + WinWinlay & &  & & & &  & \\
+ \textbf{MAC} &\textbf{93.22}  &\textbf{65.00}& &\textbf{83.34} &\textbf{55.76} & \textbf{43.31} & \textbf{42.48}\\
\toprule[1pt]
\end{tabular}
\vspace{-0.1 in}
\end{table*}
\begin{table}[t]
\caption{The results of experiments on MAC under three settings: (1)  $L_{\text{mac}}$ only includes the region-aware loss (setting \emph{R}); (2) $L_{\text{mac}}$ includes both the region-aware loss and marginal-aware loss (setting \emph{R + M}); (3) using the complete loss proposed in Eq.~\ref{loss_bacon} (setting \emph{R + M + Reg}).}
\centering	
\small
\begin{tabular}{lcccc}
\bottomrule[1pt]
\label{table4}
\multirow{2}{*}{Method}  & \multicolumn{4}{c}{HRS-Bench}\\
\cline{2-5}
 & Counting & Spatial & Size & Color \\ 
 \hline
SD1.5 &71.35 &10.80 &11.23 &13.01   \\
+ \emph{R}   &73.48 & 30.57 &14.15 &30.67 \\
+ \emph{R}  + \emph{M} & 76.03 &43.61 & 15.63 & 32.53\\
+ \emph{R}  + \emph{M}  + \emph{Reg}  & \textbf{77.88} &\textbf{ 50.39 }& \textbf{16.96} &\textbf{ 39.74} \\
\hline
SD-XL &76.99 & 34.23  & 19.96 &25.20  \\
+ \emph{R}  & 76.71 &  33.34 &30.34 &27.42 \\
+ \emph{R} + \emph{M}& 78.09  &38.82 &27.34 & 28.42\\
+ \emph{R} + \emph{M} + \emph{Reg} & \textbf{80.43} & \textbf{42.31} &\textbf{38.12} & \textbf{32.12 } \\
\hline
GLIGEN & 68.32 &44.81 &35.31 &30.55   \\
+ \emph{R}   & 79.94 & 49.43&30.87 & 35.46\\
+ \emph{R}  + \emph{M} & 82.63 & 53.08 &38.12 &40.44\\
+ \emph{R}  + \emph{M}  + \emph{Reg}  & \textbf{83.34} &\textbf{55.76} &\textbf{ 43.31} & \textbf{ 42.28} \\
\toprule[1pt]
\end{tabular}
\vspace{-0.1in}
\end{table}

\subsection{Plug and Play with MAC}
Although we mainly conduct experiments based on SD 1.5, MAC, as well as these prior training-free L2I schemes do not assume specific model architecture and can be plugged into arbitrary pre-trained Text-to-Image models using cross-attention blocks. We further evaluate the performance of MAC by deploying it on another L2I model, GLIGEN \citep{gligen}, trained by supervised learning with labeled datasets. As shown in Table~\ref{table3}, all training-free L2I schemes demonstrate improved performance across nearly all metrics quantifying controllability. Compared to SD 1.5, GLIGEN has the ability to perceive input bounding boxes and incorporate layout information into the prior conditions for sampling the initial noise $z_{0}$. With the initial latent feature $z_{0}$ generated by GLIGEN, the reverse sampler can synthesize an image with a layout that more closely aligns with the given instructions than when using $z_{0}$ produced by SD 1.5. According to Table~\ref{table3}, MAC consistently delivers the best performance in object counting, spatial relationships, and color. While R\&B outperforms MAC in the size metric, MAC achieves the second-best performance. In experiments with SD 1.5, plain SD 1.5 generates initial noise $z_{0}$ corresponding to undesirable cross-attention maps that are minimally modified through optimization, resulting in limited improvement of MAC on the size metric. In contrast, GLIGEN generates improved initial noise $z_{0}$ that facilitates MAC’s enhancement.

\subsection{Ablation Study}
To investigate the effect of $L_{\text{r}}$, $L_{\text{b}}$ and $L_{\text{reg}}$, the components of MAC’s loss / objective function, we conduct additional experiments using SD 1.5, SD-XL, and GLIGEN under three settings: (1) only the region-attention loss $L_{\text{r}}$ is used (setting \emph{R}); (2) $L_{\text{r}}$ is combined with the marginal-attention loss $L_{\text{m}}$ (setting \emph{R + M}); (3) using the complete loss objective as described in Eq.~\ref{loss_bacon} (setting \emph{R + M + Reg}). 

\begin{figure}[t] 
    \centering
	  \subfloat[\emph{R}]{
    \includegraphics[width=0.33\linewidth]{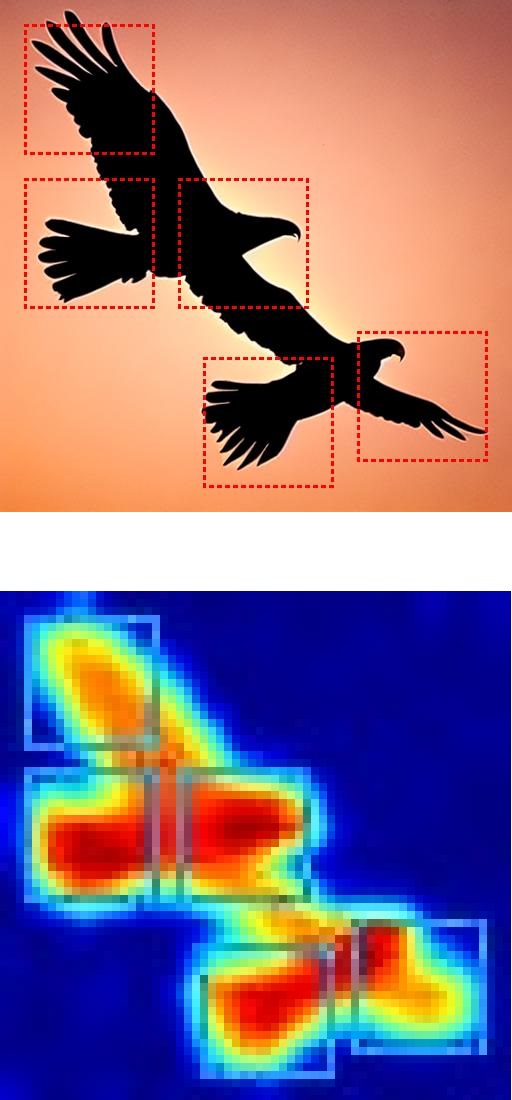}}
	  \subfloat[\emph{R + M}]{
        \includegraphics[width=0.33\linewidth]{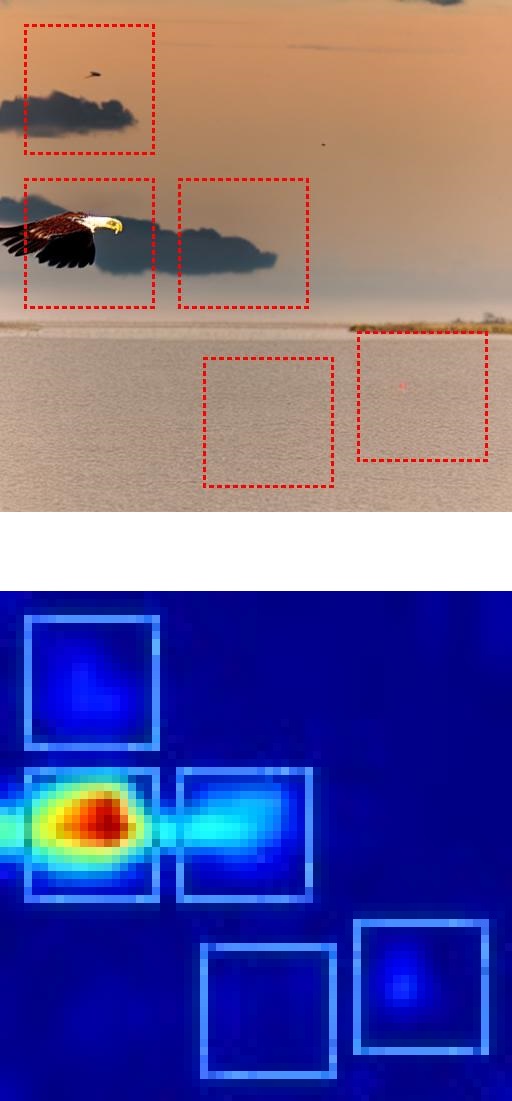}}
	  \subfloat[\emph{R + M + Reg}]{
        \includegraphics[width=0.33\linewidth]{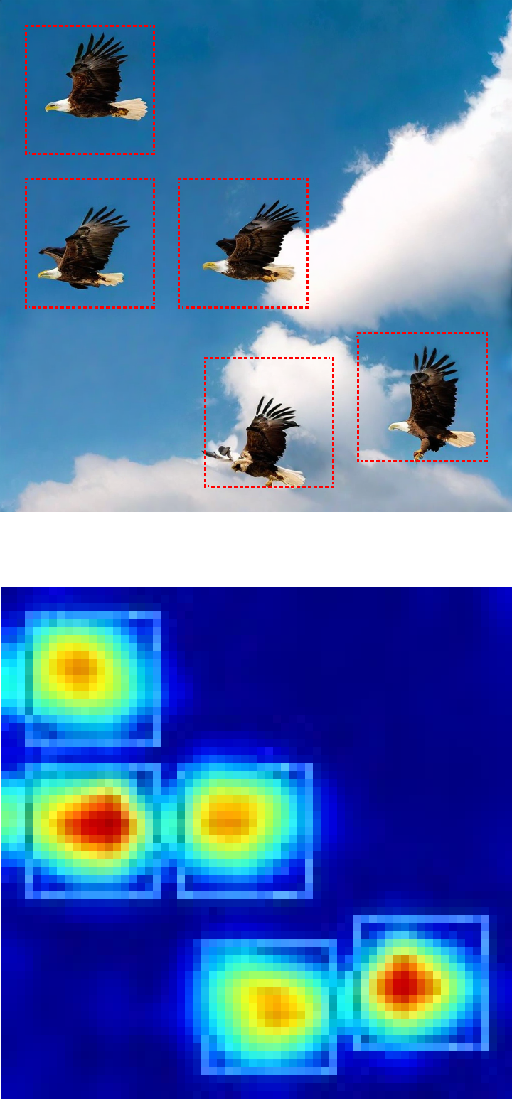}}
        
	\caption{Visualization of cross-attention maps of targeting objects generated in three settings: (1) \emph{R}; (2) \emph{R + M}; (3) \emph{R + M +Reg}.}
\label{abltion_study} 
\end{figure}

When using only the region-attention loss, MAC synthesizes images similar to those produced by R\&B, with the cross-attention map exhibiting the same overlapping issue discussed earlier -- see the illustration in Figure~\ref{abltion_study}(a). Adding the marginal-attention loss to the objective resolves the overlapping problem but may cause attention vanishing, as shown in Figure~\ref{abltion_study}(b). This issue does not occur when there is only one bounding box per phrase, due to the constraint imposed by $L_{\text{r}}$. Moreover, the regularization loss $L_{\text{reg}}$ addresses the attention vanishing problem, ensuring that each bounding box contains exactly one object in the generated images, as shown in Figure~\ref{abltion_study}(c). 

The quantitative results in Table~\ref{table4}, obtained in experiments on three distinct pretrained diffusion models, further demonstrate the improvement achieved by including $L_{\text{m}}$ and $L_{\text{reg}}$ in the objective function. However, the attention vanishing problem typically arises in settings where multiple objects are assigned adjacent bounding boxes. The text prompts in HRS are relatively simple, so there is minimal difference between \emph{R + M} and \emph{R + M + Reg} settings.

\section{Conclusion}
We proposed MAC, a novel training-free layout-to-image (L2I) generation scheme utilizing pre-trained diffusion models. We identify two inherent issues that lead to semantic errors and incorrect object counts in generation: (1) coarse-grained cross-attention maps, and (2) overlapping cross-attention. To address these issues, we leverage self-attention maps to refine cross-attention maps and design an objective that combines three distinct loss functions. %to mitigate these issues. 
Comprehensive quantitative and visual results demonstrate that MAC outperforms state-of-the-art L2I schemes and consistently adapts well across varying model architectures. We envision our MAC as a milestone that can inspire the follow-up research in training-free layout-to-image generation.
{
    \small
    \bibliographystyle{ieeenat_fullname}
    \bibliography{main}

\begin{thebibliography}{37}
\providecommand{\natexlab}[1]{#1}
\providecommand{\url}[1]{\texttt{#1}}
\expandafter\ifx\csname urlstyle\endcsname\relax
  \providecommand{\doi}[1]{doi: #1}\else
  \providecommand{\doi}{doi: \begingroup \urlstyle{rm}\Url}\fi

\bibitem[Avrahami et~al.(2023)Avrahami, Hayes, Gafni, Gupta, Taigman, Parikh, Lischinski, Fried, and Yin]{avrahami2023spatext}
Omri Avrahami, Thomas Hayes, Oran Gafni, Sonal Gupta, Yaniv Taigman, Devi Parikh, Dani Lischinski, Ohad Fried, and Xi Yin.
\newblock Spatext: Spatio-textual representation for controllable image generation.
\newblock In \emph{Proceedings of the IEEE/CVF Conference on Computer Vision and Pattern Recognition}, pages 18370--18380, 2023.

\bibitem[Bakr et~al.(2023)Bakr, Sun, Shen, Khan, Li, and Elhoseiny]{hrs}
Eslam~Mohamed Bakr, Pengzhan Sun, Xiaogian Shen, Faizan~Farooq Khan, Li~Erran Li, and Mohamed Elhoseiny.
\newblock Hrs-bench: Holistic, reliable and scalable benchmark for text-to-image models.
\newblock In \emph{Proceedings of the IEEE/CVF International Conference on Computer Vision}, pages 20041--20053, 2023.

\bibitem[Chefer et~al.(2023)Chefer, Alaluf, Vinker, Wolf, and Cohen-Or]{attend-and-excite}
Hila Chefer, Yuval Alaluf, Yael Vinker, Lior Wolf, and Daniel Cohen-Or.
\newblock Attend-and-excite: Attention-based semantic guidance for text-to-image diffusion models.
\newblock \emph{ACM Transactions on Graphics (TOG)}, 42\penalty0 (4):\penalty0 1--10, 2023.

\bibitem[Chen et~al.(2023)Chen, Xie, Chen, Wang, Hong, Li, and Yeung]{geodiffusion}
Kai Chen, Enze Xie, Zhe Chen, Yibo Wang, Lanqing Hong, Zhenguo Li, and Dit-Yan Yeung.
\newblock Geodiffusion: Text-prompted geometric control for object detection data generation.
\newblock \emph{arXiv preprint arXiv:2306.04607}, 2023.

\bibitem[Chen et~al.(2024)Chen, Laina, and Vedaldi]{layout_guidance}
Minghao Chen, Iro Laina, and Andrea Vedaldi.
\newblock Training-free layout control with cross-attention guidance.
\newblock In \emph{Proceedings of the IEEE/CVF Winter Conference on Applications of Computer Vision}, pages 5343--5353, 2024.

\bibitem[Cheng et~al.(2024)Cheng, Ma, Wu, Liu, Ma, Wu, Leng, and Yin]{cheng2024hico}
Bo Cheng, Yuhang Ma, Liebucha Wu, Shanyuan Liu, Ao Ma, Xiaoyu Wu, Dawei Leng, and Yuhui Yin.
\newblock Hico: Hierarchical controllable diffusion model for layout-to-image generation.
\newblock \emph{arXiv preprint arXiv:2410.14324}, 2024.

\bibitem[Goodfellow et~al.(2020)Goodfellow, Pouget-Abadie, Mirza, Xu, Warde-Farley, Ozair, Courville, and Bengio]{gan}
Ian Goodfellow, Jean Pouget-Abadie, Mehdi Mirza, Bing Xu, David Warde-Farley, Sherjil Ozair, Aaron Courville, and Yoshua Bengio.
\newblock Generative adversarial networks.
\newblock \emph{Communications of the ACM}, 63\penalty0 (11):\penalty0 139--144, 2020.

\bibitem[Hertz et~al.(2022)Hertz, Mokady, Tenenbaum, Aberman, Pritch, and Cohen-Or]{prompt}
Amir Hertz, Ron Mokady, Jay Tenenbaum, Kfir Aberman, Yael Pritch, and Daniel Cohen-Or.
\newblock Prompt-to-prompt image editing with cross attention control.
\newblock \emph{arXiv preprint arXiv:2208.01626}, 2022.

\bibitem[Ho et~al.(2020)Ho, Jain, and Abbeel]{ddpm}
Jonathan Ho, Ajay Jain, and Pieter Abbeel.
\newblock Denoising diffusion probabilistic models.
\newblock \emph{Advances in neural information processing systems}, 33:\penalty0 6840--6851, 2020.

\bibitem[Kanopoulos et~al.(1988{\natexlab{a}})Kanopoulos, Vasanthavada, and Baker]{edge}
Nick Kanopoulos, Nagesh Vasanthavada, and Robert~L Baker.
\newblock Design of an image edge detection filter using the sobel operator.
\newblock \emph{IEEE Journal of solid-state circuits}, 23\penalty0 (2):\penalty0 358--367, 1988{\natexlab{a}}.

\bibitem[Kanopoulos et~al.(1988{\natexlab{b}})Kanopoulos, Vasanthavada, and Baker]{sobel}
Nick Kanopoulos, Nagesh Vasanthavada, and Robert~L Baker.
\newblock Design of an image edge detection filter using the sobel operator.
\newblock \emph{IEEE Journal of solid-state circuits}, 23\penalty0 (2):\penalty0 358--367, 1988{\natexlab{b}}.

\bibitem[Kim et~al.(2023)Kim, Lee, Kim, Ha, and Zhu]{dense}
Yunji Kim, Jiyoung Lee, Jin-Hwa Kim, Jung-Woo Ha, and Jun-Yan Zhu.
\newblock Dense text-to-image generation with attention modulation.
\newblock In \emph{Proceedings of the IEEE/CVF International Conference on Computer Vision}, pages 7701--7711, 2023.

\bibitem[Li et~al.(2025)Li, Hu, Liu, and Wang]{li2025winwinlay}
Bonan Li, Yinhan Hu, Songhua Liu, and Xinchao Wang.
\newblock Control and realism: Best of both worlds in layout-to-image without training.
\newblock \emph{arXiv preprint arXiv:2506.15563}, 2025.

\bibitem[Li et~al.(2023)Li, Liu, et~al.]{gligen}
Yuheng Li, Haotian Liu, et~al.
\newblock Gligen: Open-set grounded text-to-image generation.
\newblock \emph{CVPR}, 2023.

\bibitem[Lin et~al.(2014)Lin, Maire, Belongie, Hays, Perona, Ramanan, Doll{\'a}r, and Zitnick]{coco}
Tsung-Yi Lin, Michael Maire, Serge Belongie, James Hays, Pietro Perona, Deva Ramanan, Piotr Doll{\'a}r, and C~Lawrence Zitnick.
\newblock Microsoft coco: Common objects in context.
\newblock In \emph{Computer Vision--ECCV 2014: 13th European Conference, Zurich, Switzerland, September 6-12, 2014, Proceedings, Part V 13}, pages 740--755. Springer, 2014.

\bibitem[Liu et~al.(2024)Liu, Huang, and Xu]{liu2024csg}
Jiaqi Liu, Tao Huang, and Chang Xu.
\newblock Training-free composite scene generation for layout-to-image synthesis, 2024.
\newblock \emph{URL https://arxiv. org/abs/2407.13609}, 2024.

\bibitem[Liu et~al.(2023)Liu, Zeng, Ren, Li, Zhang, Yang, Li, Yang, Su, Zhu, et~al.]{grounddino}
Shilong Liu, Zhaoyang Zeng, Tianhe Ren, Feng Li, Hao Zhang, Jie Yang, Chunyuan Li, Jianwei Yang, Hang Su, Jun Zhu, et~al.
\newblock Grounding dino: Marrying dino with grounded pre-training for open-set object detection.
\newblock \emph{arXiv preprint arXiv:2303.05499}, 2023.

\bibitem[Ma et~al.(2023)Ma, Yang, Ju, Zhang, Liu, Wang, Zhang, and Wang]{ma2023diffusionseg}
Chaofan Ma, Yuhuan Yang, Chen Ju, Fei Zhang, Jinxiang Liu, Yu Wang, Ya Zhang, and Yanfeng Wang.
\newblock Diffusionseg: Adapting diffusion towards unsupervised object discovery.
\newblock \emph{arXiv preprint arXiv:2303.09813}, 2023.

\bibitem[Nguyen et~al.(2024)Nguyen, Vu, Tran, and Nguyen]{nguyen2024dataset}
Quang Nguyen, Truong Vu, Anh Tran, and Khoi Nguyen.
\newblock Dataset diffusion: Diffusion-based synthetic data generation for pixel-level semantic segmentation.
\newblock \emph{Advances in Neural Information Processing Systems}, 36, 2024.

\bibitem[Phung et~al.(2024)Phung, Ge, and Huang]{anr}
Quynh Phung, Songwei Ge, and Jia-Bin Huang.
\newblock Grounded text-to-image synthesis with attention refocusing.
\newblock In \emph{Proceedings of the IEEE/CVF Conference on Computer Vision and Pattern Recognition}, pages 7932--7942, 2024.

\bibitem[Podell et~al.(2023)Podell, English, Lacey, Blattmann, Dockhorn, M{\"u}ller, Penna, and Rombach]{sdxl}
Dustin Podell, Zion English, Kyle Lacey, Andreas Blattmann, Tim Dockhorn, Jonas M{\"u}ller, Joe Penna, and Robin Rombach.
\newblock Sdxl: Improving latent diffusion models for high-resolution image synthesis.
\newblock \emph{arXiv preprint arXiv:2307.01952}, 2023.

\bibitem[Radford et~al.(2021)Radford, Kim, Hallacy, Ramesh, Goh, Agarwal, Sastry, Askell, Mishkin, Clark, et~al.]{clip}
Alec Radford, Jong~Wook Kim, Chris Hallacy, Aditya Ramesh, Gabriel Goh, Sandhini Agarwal, Girish Sastry, Amanda Askell, Pamela Mishkin, Jack Clark, et~al.
\newblock Learning transferable visual models from natural language supervision.
\newblock In \emph{International conference on machine learning}, pages 8748--8763. PMLR, 2021.

\bibitem[Ramesh et~al.(2022)Ramesh, Dhariwal, Nichol, Chu, and Chen]{Dall-E}
Aditya Ramesh, Prafulla Dhariwal, Alex Nichol, Casey Chu, and Mark Chen.
\newblock Hierarchical text-conditional image generation with clip latents.
\newblock \emph{arXiv preprint arXiv:2204.06125}, 1\penalty0 (2):\penalty0 3, 2022.

\bibitem[Rombach et~al.(2022)Rombach, Blattmann, Lorenz, Esser, and Ommer]{LDM}
Robin Rombach, Andreas Blattmann, Dominik Lorenz, Patrick Esser, and Bj{\"o}rn Ommer.
\newblock High-resolution image synthesis with latent diffusion models.
\newblock In \emph{Proceedings of the IEEE/CVF conference on computer vision and pattern recognition}, pages 10684--10695, 2022.

\bibitem[Ronneberger et~al.(2015)Ronneberger, Fischer, and Brox]{unet}
Olaf Ronneberger, Philipp Fischer, and Thomas Brox.
\newblock U-net: Convolutional networks for biomedical image segmentation.
\newblock In \emph{Medical image computing and computer-assisted intervention--MICCAI 2015: 18th international conference, Munich, Germany, October 5-9, 2015, proceedings, part III 18}, pages 234--241. Springer, 2015.

\bibitem[Saharia et~al.(2022)Saharia, Chan, Saxena, Li, Whang, Denton, Ghasemipour, Gontijo~Lopes, Karagol~Ayan, Salimans, et~al.]{Imagen}
Chitwan Saharia, William Chan, Saurabh Saxena, Lala Li, Jay Whang, Emily~L Denton, Kamyar Ghasemipour, Raphael Gontijo~Lopes, Burcu Karagol~Ayan, Tim Salimans, et~al.
\newblock Photorealistic text-to-image diffusion models with deep language understanding.
\newblock \emph{Advances in neural information processing systems}, 35:\penalty0 36479--36494, 2022.

\bibitem[Schuhmann et~al.(2022)Schuhmann, Beaumont, Vencu, Gordon, Wightman, Cherti, Coombes, Katta, Mullis, Wortsman, et~al.]{laion}
Christoph Schuhmann, Romain Beaumont, Richard Vencu, Cade Gordon, Ross Wightman, Mehdi Cherti, Theo Coombes, Aarush Katta, Clayton Mullis, Mitchell Wortsman, et~al.
\newblock Laion-5b: An open large-scale dataset for training next generation image-text models.
\newblock \emph{Advances in Neural Information Processing Systems}, 35:\penalty0 25278--25294, 2022.

\bibitem[Song et~al.(2020)Song, Meng, and Ermon]{ddim}
Jiaming Song, Chenlin Meng, and Stefano Ermon.
\newblock Denoising diffusion implicit models.
\newblock \emph{arXiv preprint arXiv:2010.02502}, 2020.

\bibitem[Song et~al.(2021)Song, Sohl-Dickstein, Kingma, Kumar, Ermon, and Poole]{sde}
Yang Song, Jascha Sohl-Dickstein, Diederik~P Kingma, Abhishek Kumar, Stefano Ermon, and Ben Poole.
\newblock Score-based generative modeling through stochastic differential equations.
\newblock In \emph{International Conference on Learning Representations}, 2021.

\bibitem[Taghipour et~al.(2024)Taghipour, Ghahremani, Bennamoun, Rekavandi, Laga, and Boussaid]{taghipour2024box}
Ashkan Taghipour, Morteza Ghahremani, Mohammed Bennamoun, Aref~Miri Rekavandi, Hamid Laga, and Farid Boussaid.
\newblock Box it to bind it: Unified layout control and attribute binding in t2i diffusion models.
\newblock \emph{arXiv preprint arXiv:2402.17910}, 2024.

\bibitem[Xiao et~al.(2023)Xiao, Li, et~al.]{rnb}
Jiayu Xiao, Liang Li, et~al.
\newblock R\&b: Region and boundary aware zero-shot grounded text-to-image generation.
\newblock \emph{ICLR}, 2023.

\bibitem[Xie et~al.(2023)Xie, Li, Huang, Liu, Zhang, Zheng, and Shou]{boxdiff}
Jinheng Xie, Yuexiang Li, Yawen Huang, Haozhe Liu, Wentian Zhang, Yefeng Zheng, and Mike~Zheng Shou.
\newblock Boxdiff: Text-to-image synthesis with training-free box-constrained diffusion.
\newblock In \emph{Proceedings of the IEEE/CVF International Conference on Computer Vision}, pages 7452--7461, 2023.

\bibitem[Yang et~al.(2023)Yang, Wang, Gan, Li, Lin, Wu, Duan, Liu, Liu, Zeng, et~al.]{reco}
Zhengyuan Yang, Jianfeng Wang, Zhe Gan, Linjie Li, Kevin Lin, Chenfei Wu, Nan Duan, Zicheng Liu, Ce Liu, Michael Zeng, et~al.
\newblock Reco: Region-controlled text-to-image generation.
\newblock In \emph{Proceedings of the IEEE/CVF Conference on Computer Vision and Pattern Recognition}, pages 14246--14255, 2023.

\bibitem[Zhang et~al.(2025)Zhang, Hong, Wang, Shao, Wu, Wu, and Jiang]{zhang2025creatilayout}
Hui Zhang, Dexiang Hong, Yitong Wang, Jie Shao, Xinglong Wu, Zuxuan Wu, and Yu-Gang Jiang.
\newblock Creatilayout: Siamese multimodal diffusion transformer for creative layout-to-image generation.
\newblock In \emph{Proceedings of the IEEE/CVF International Conference on Computer Vision}, pages 18487--18497, 2025.

\bibitem[Zhang et~al.(2023)Zhang, Rao, and Agrawala]{controlnet}
Lvmin Zhang, Anyi Rao, and Maneesh Agrawala.
\newblock Adding conditional control to text-to-image diffusion models.
\newblock In \emph{Proceedings of the IEEE/CVF International Conference on Computer Vision}, pages 3836--3847, 2023.

\bibitem[Zhao et~al.(2023)Zhao, Li, Jin, and Zhou]{loco}
Peiang Zhao, Han Li, Ruiyang Jin, and S~Kevin Zhou.
\newblock Loco: Locally constrained training-free layout-to-image synthesis.
\newblock \emph{arXiv preprint arXiv:2311.12342}, 2023.

\bibitem[Zheng et~al.(2023)Zheng, Zhou, Li, Qi, Shan, and Li]{layout}
Guangcong Zheng, Xianpan Zhou, Xuewei Li, Zhongang Qi, Ying Shan, and Xi Li.
\newblock Layoutdiffusion: Controllable diffusion model for layout-to-image generation.
\newblock In \emph{Proceedings of the IEEE/CVF Conference on Computer Vision and Pattern Recognition}, pages 22490--22499, 2023.

\end{thebibliography}
}

% WARNING: do not forget to delete the supplementary pages from your submission 
\end{document}

\maketitle

\section*{Appendix A: Experimental Details}
\label{exp_detail}
In this section, we report the model architectures and hyper-parameters used in the experiments across all layout-to-image schemes. We employed the DDIM scheduler \citep{ddim} with $50$ denoising steps in all experiments, performing latent variable optimization only within the first 10 steps, with a maximum of 5 iterations per step. The step size for updating the latent variable $z_{t}$ was set to $70$, and the loss threshold value for early stopping was set to $10^{-6}$. In all experiments, the weight for classifier-free guidance was set to $7.5$. In the experiments with MAC, $\lambda$ and $\alpha$ for combining boundary-attention loss and regularization loss were both set to $0.25$.

\section*{Appendix B: Evaluation Details}
\label{metric}
We employed the state-of-the-art GroundDINO \citep{grounddino} to detect objects in the synthetic images generated with the input prompts and layout instruction from DrawBench and HRS benchmarks. Specifically, GroundDINO generates multiple predicted bounding boxes corresponding to predicted categories on the synthetic images with threshold value for confidence $0.25$. Those boxes are then used to compute various metrics for measuring the spatial controllability and counting accuracy.

\subsection*{Object Counting}
To evaluate the layout-to-image schemes in object counting, we record the number of predicted bounding boxes $n^{(i)}_{\text{pred}}$ corresponding to the phrase $\mathbf{p}^{(i)}$, and compute the correct number of boxes and false number of boxes by
\begin{equation}
    n^{(i)}_{\text{cor}} = \min(n_{\text{pred}}^{(i)}, n^{(i)}_{\text{gt}}),
\end{equation}
\begin{equation}
    n^{(i)}_{\text{fal}} = \max(  n_{\text{pred}}^{(i)} - n^{(i)}_{\text{gt}}, 0),
\end{equation}
\begin{equation}
    n^{(i)}_{\text{neg}} = \max(n^{(i)}_{\text{gt}} - n_{\text{pred}}^{(i)}, 0),
\end{equation}
where $n^{(i)}_{\text{gt}}$ is the ground-truth number. With $n^{(i)}_{\text{cor}}$ and $n^{(i)}_{\text{fal}}$, we can obtain \begin{equation}
    \emph{precision} = \frac{\sum_{i\in \mathcal{I}}n^{(i)}_{\text{cor}}}{\sum_{i\in \mathcal{I}}n^{(i)}_{\text{cor}} + \sum_{i\in \mathcal{I}}n^{(i)}_{\text{fal}}},
\end{equation}
\begin{equation}
    \emph{recall} = \frac{\sum_{i\in \mathcal{I}}n^{(i)}_{\text{cor}}}{\sum_{i\in \mathcal{I}}n^{(i)}_{\text{cor}} + \sum_{i\in \mathcal{I}}n^{(i)}_{\text{neg}}},
\end{equation}
where $\mathcal{I}$ denotes indices of the phrases in the prompt, which is defined in the Section 3.1 in the main paper. With the metrics of precision and recall, we compute \emph{F1} score by
\begin{equation}
    \emph{F1} = \frac{2*\emph{precision}*\emph{recall}}{\emph{precision} + \emph{recall}}.
\end{equation}

\subsection*{Spatial Accuracy}
In the experiments on spatial relationship, phrases in the input prompts are given grouth-truth relationship. For instance, in the prompt \emph{a cat is on the left of a dog}, the two phrases \emph{cat} and \emph{dog} have the spatial relationship \emph{on the left of}. By comparing the mean points of the predicted boxes for \emph{cat} and \emph{dog} as
\begin{equation}
    (\mathbf{x}_{1}, \mathbf{y}_{1}) = (\frac{\mathbf{x}_{1}^{\text{left}} + \mathbf{x}_{1}^{\text{right}}}{2},\frac{\mathbf{y}_{1}^{\text{left}} + \mathbf{y}_{1}^{\text{right}}}{2}),
\end{equation}
\begin{equation}
    (\mathbf{x}_{2}, \mathbf{y}_{2}) = (\frac{\mathbf{x}_{2}^{\text{left}} + \mathbf{x}_{2}^{\text{right}}}{2},\frac{\mathbf{y}_{2}^{\text{left}} + \mathbf{y}_{2}^{\text{right}}}{2}),
\end{equation}
and record the number of correct prediction
\begin{equation}
 n_{\text{cor}}^{(i)}= \left\{
    \begin{aligned}
        & \; 1, \text{ if } (\mathbf{x}_{1}, \mathbf{y}_{1})  \text{ is on the \emph{ground-truth} of } (\mathbf{x}_{2}, \mathbf{y}_{2}), \\
        & \; 0, \text{ if } (\mathbf{x}_{1}, \mathbf{y}_{1})  \text{ is not on the \emph{ground-truth} of } (\mathbf{x}_{2}, \mathbf{y}_{2}),\\
    \end{aligned}
    \right.
\end{equation}
and compute the spatial accuracy
\begin{equation}
    \emph{ACC}_{\text{spatial}} = \frac{\sum_{i\in \mathcal{I}}n_{\text{cor}}^{(i)}}{|\mathcal{I}|}.
\end{equation}

\subsection*{Size Accuracy}
In the experiments on size relationship, phrases in the input prompts are given grouth-truth relationship. For instance, in the prompt \emph{a cat is smaller than a dog}, the two phrases \emph{cat} and \emph{dog} have the size relationship \emph{smaller}. By comparing the area of the predicted boxes $\mathcal{B}_{1}$ and $\mathcal{B}_{2}$ for \emph{cat} and \emph{dog} as
\begin{equation}
 n_{\text{cor}}^{(i)}= \left\{
    \begin{aligned}
        & \; 1, \text{ if } \text{Area}(\mathcal{B}_{1})  \text{ is \emph{ground-truth} than } \text{Area}(\mathcal{B}_{2}), \\
        & \; 0, \text{ if } \text{Area}(\mathcal{B}_{1})  \text{ is not \emph{ground-truth} than } \text{Area}(\mathcal{B}_{2}),\\
    \end{aligned}
    \right.
\end{equation}
and compute the size accuracy
\begin{equation}
    \emph{ACC}_{\text{size}} = \frac{\sum_{i\in \mathcal{I}}n_{\text{cor}}^{(i)}}{|\mathcal{I}|}.
\end{equation}

\begin{figure}[t] 
    \centering
	  \subfloat[one penguin]{
       \includegraphics[width=0.3\linewidth]
       {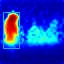}}
	  \subfloat[baby penguins]{
        \includegraphics[width=0.3\linewidth]{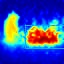}}
        \subfloat[whole image]{
        \includegraphics[width=0.3\linewidth]{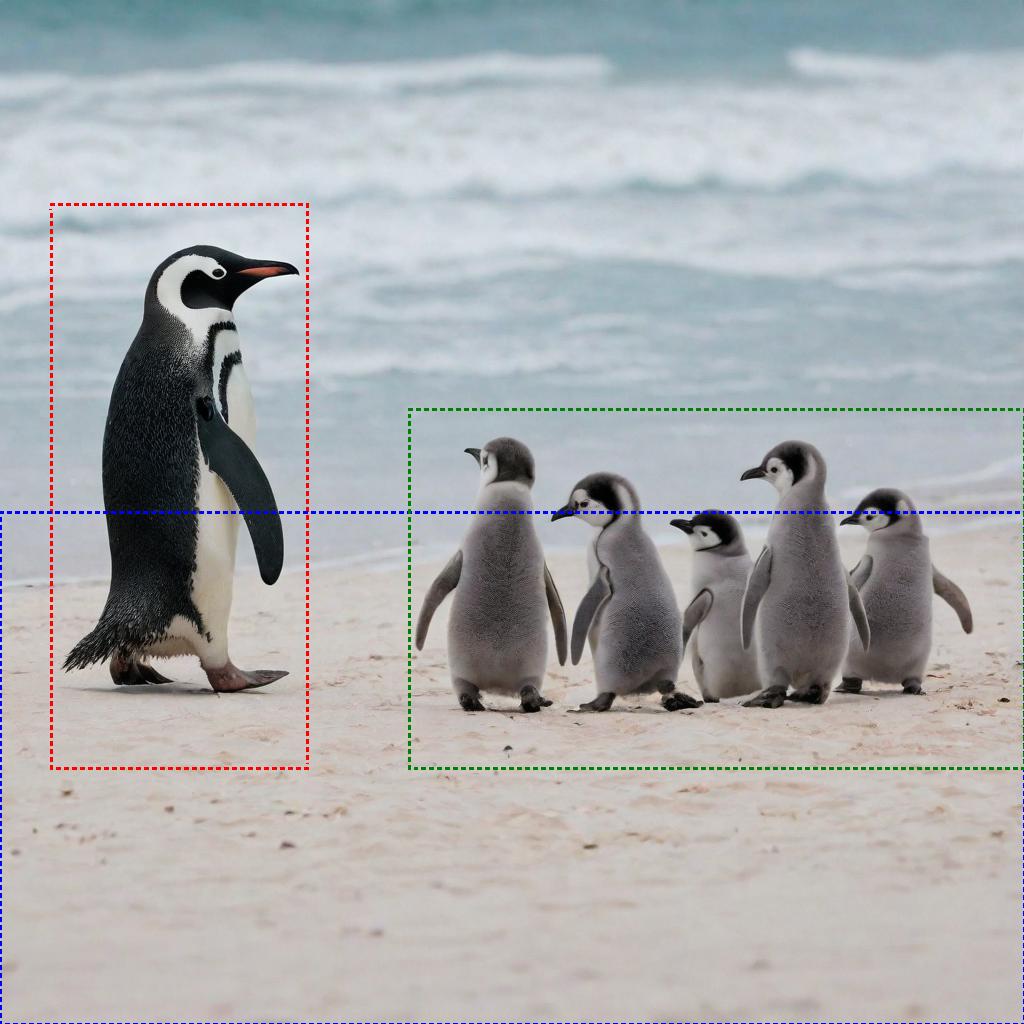}}
	\caption{Example of overlapping objects in a box.}
\label{fig1} 
\end{figure}

\subsection*{Color Accuracy}
In the experiments on color, phrases in the input prompts are given color instruction. For instance, in the prompt \emph{a white cat and a black dog}, the two phrases \emph{cat} and \emph{dog} have the color \emph{white} and \emph{black}. Similarly, we can compute the color accuracy:
\begin{equation}
 n_{\text{cor}}^{(i)}= \left\{
    \begin{aligned}
        & \; 1, \text{ if } \text{ the predicted color is \emph{ground-truth}, }  \\
        & \; 0, \text{ if } \text{ the predicted color is not \emph{ground-truth}, } \\
    \end{aligned}
    \right.
\end{equation}
and compute the color accuracy
\begin{equation}
    \emph{ACC}_{\text{color}} = \frac{\sum_{i\in \mathcal{I}}n_{\text{cor}}^{(i)}}{|\mathcal{I}|}.
\end{equation}

\subsection*{Complex Layouts}
As shown in Figure~\ref{fig1}, our method excels at handling complex and crowded scenes compared to previous approaches, thanks to the boundary-attention constraint. For \textbf{overlapping objects}, the user can assign a unique bounding box to each object and use a distinct phrase to describe them. For example: ``One penguin is standing on the left of the beach, and five baby penguins are on the right." The user can assign one box for the phrase \textcolor{blue}{One penguin} while one box for the phrase \textcolor{orange}{baby penguins}, as illustrated on the top right. However, the cross-attention maps for multiple objects within the same box inevitably overlap, leading to incorrect counting. We recognize that these training-free L2I schemes, including ours, may struggle with accurate counting in overlapping boxes, resulting in unsatisfactory generation in crowded scenarios. Addressing this limitation would be interesting for future work.

\subsection*{Hyper-parameter Sensitivity}
The results in Table 4 highlight the roles of the three loss functions: (1) region-attention loss regulates object location, (2) boundary-attention loss improves object counting accuracy, and (3) regularization loss prevents the attention map from vanishing. Therefore, we set $\lambda = \alpha = 0.5 < 1 $ as location regulation is the highest priority. To provide further insights, we will conduct additional experiments with varying $\lambda $ and $\alpha$ in the revised version.

\subsection*{Text-Layout Mismatch}
 The entire discussion on improving object counting is based on the assumption that the user inputs a consistent layout and text prompts, which is a reasonable assumption, our method will not improve the counting otherwise. Under the same assumption, the existing schemes have shown poor performance in object counting accuracy, as illustrated in Table 1, because it is challenging to accurately determine the count relying solely on the text prompt.

{
    \small
    \bibliographystyle{ieeenat_fullname}
    \bibliography{main}
}